\documentclass[10pt,journal,compsoc]{IEEEtran}
\ifCLASSOPTIONcompsoc
  \usepackage[nocompress]{cite}
\else
  \usepackage{cite}
\fi
\ifCLASSINFOpdf
\else
\fi
\usepackage{url}

\usepackage{graphicx}
\usepackage{amsmath}
\usepackage{amssymb}
\usepackage{booktabs}

\usepackage{makecell}
\usepackage{multirow}
\usepackage{booktabs}
\usepackage{amsmath}
\usepackage{amssymb}
\usepackage{diagbox}
\usepackage{float}
\usepackage{color}
\usepackage{xcolor}
\usepackage{tabu}

\def\onedot{.}

\def\etal{\emph{et al}\onedot}

\newcommand{\revise}{black}

\begin{document}
%
\title{Vision Transformer with Attentive Pooling for \\ Robust Facial Expression Recognition}
%
%
%
%

\author{Fanglei~Xue,
        Qiangchang~Wang,
        Zichang~Tan,
        Zhongsong~Ma,
        and~Guodong~Guo,~\IEEEmembership{Senior Member,~IEEE}
\IEEEcompsocitemizethanks{
\IEEEcompsocthanksitem This work was done when Fanglei Xue and Qiangchang Wang were interns at Institute of Deep Learning, Baidu Research. The first two authors contribute equally. \textit{Corresponding authors: Zhongsong Ma and Guodong Guo.}
\IEEEcompsocthanksitem Fanglei Xue and Zhongsong Ma are with University of Chinese Academy of Sciences, and also with the Key Laboratory of Space Utilization, Technology and Engineering Center for Space Utilization, Chinese Academy of Sciences, Beijing, China. (xuefanglei19@mails.ucas.ac.cn; mazhongsong@csu.ac.cn)
\IEEEcompsocthanksitem Qiangchang Wang is with West Virginia University, Morgantown, USA. (qw0007@mix.wvu.edu)
\IEEEcompsocthanksitem Zichang Tan and Guodong Guo are with Institute of Deep Learning, Baidu Research, and also with National Engineering Laboratory for Deep Learning Technology and Application, Beijing, China (tanzichang@baidu.com; guodong.guo@mail.wvu.edu)
}

\thanks{Manuscript received April 19, 2005; revised August 26, 2015.}}

%
%

\markboth{Journal of \LaTeX\ Class Files,~Vol.~14, No.~8, August~2015}%
{Shell \MakeLowercase{\textit{et al.}}: Bare Demo of IEEEtran.cls for Computer Society Journals}
%



\IEEEtitleabstractindextext{%
\begin{abstract}
Facial Expression Recognition (FER) in the wild is an extremely challenging task. Recently, some Vision Transformers (ViT) have been explored for FER, but most of them perform inferiorly compared to Convolutional Neural Networks (CNN). This is mainly because the new proposed modules are difficult to converge well from scratch due to lacking inductive bias and easy to focus on the occlusion and noisy areas. TransFER, a representative transformer-based method for FER, alleviates this with multi-branch attention dropping but brings excessive computations. On the contrary, we present two attentive pooling (AP) modules to pool noisy features directly.
The AP modules include Attentive Patch Pooling (APP) and Attentive Token Pooling (ATP). They aim to guide the model to emphasize the most discriminative features while reducing the impacts of less relevant features.
The proposed APP is employed to select the most informative patches on CNN features, and ATP discards unimportant tokens in ViT. Being simple to implement and without learnable parameters, the APP and ATP intuitively reduce the computational cost while boosting the performance by ONLY pursuing the most discriminative features. Qualitative results demonstrate the motivations and effectiveness of our attentive poolings. Besides, quantitative results on six in-the-wild datasets outperform other state-of-the-art methods.
\end{abstract}

}

\maketitle

\IEEEdisplaynontitleabstractindextext

\ifCLASSOPTIONpeerreview
\begin{center} \bfseries EDICS Category: 3-BBND \end{center}
\fi
%
\IEEEpeerreviewmaketitle

\IEEEraisesectionheading{\section{Introduction}\label{sec:introduction}}

\IEEEPARstart{F}{acial}
expression is one of the most important ways for people to express their emotions \cite{darwin1998expression}. Facial Expression Recognition (FER) requires that the computer program could automatically recognize the expression from an input face image. The FER task has attracted broad interest in the computer vision community \cite{shan2009facial, mollahosseini2017affectnet, li2017reliable} due to its wide applications in human-computer interactions, medical progress monitoring, driver fatigue monitoring, and many other fields.

\textcolor{\revise}{
However, FER is a very challenging task, especially in the wild. This is mainly because of the significant intra-class variances and inter-class similarities among expression categories, which differ from the general image classification task. For example, the same people in the same illumination and pose may have different expressions, while people with different identities, ages, gender, and pose may express the same emotion. In the past few years, with the development of the convolutional neural network, many methods~\cite{li2017reliable,cai2018IslandLoss,Li2019OcclusionAware,zeng2018facial,wang2020region,fanFacialExpressionRecognition2020} have been proposed and greatly improve the performance of FER.
}

\begin{figure}[t]
\begin{center}
\includegraphics[width=7.6cm]{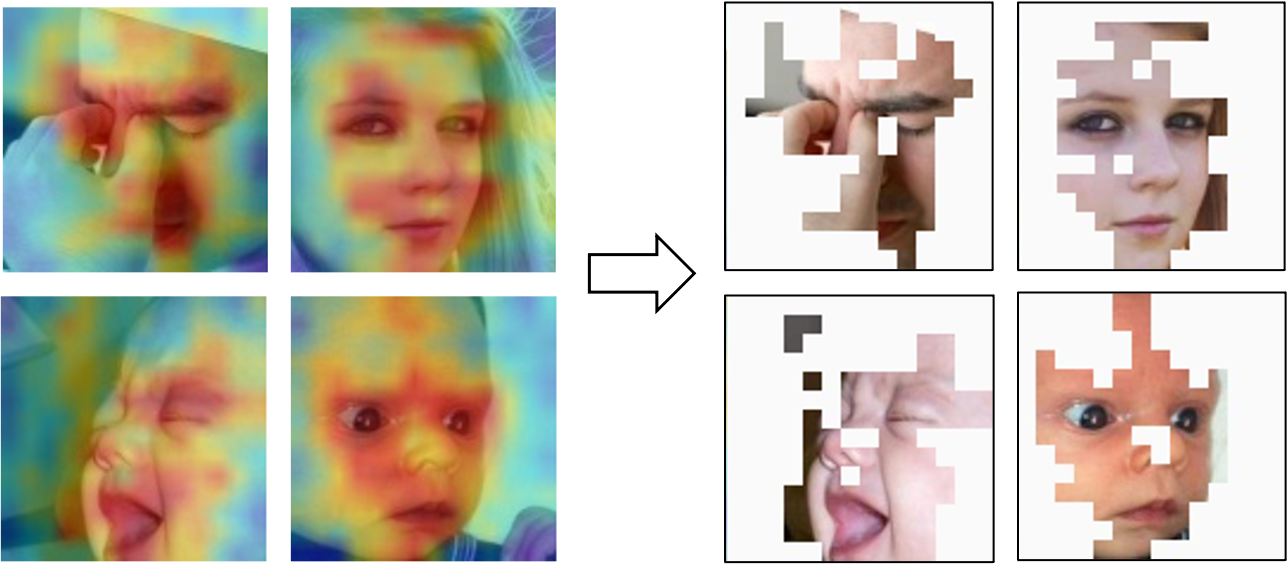}
\end{center}
\caption{An illustration of attention maps in TransFER \cite{xue2021TransFER} (left) and the attentive pooling results of APP (right). Our APP directly discards the background features to reduce the influence of noisy areas and save computation time. }
\label{fig:la_attn}
\end{figure}

Recently, the Vision Transformer (ViT) was proposed for image classification \cite{dosovitskiy2020image} and achieved promising performance with the non-local self-attention mechanism. It shows great potential for solving visual tasks. Some researchers adopt the ViT for FER \cite{ma2021RobustFacial,li2021MViTMask,xue2021TransFER}.
However, the performance is inferior to the state-of-the-art CNNs except TransFER \cite{xue2021TransFER}. 
\textcolor{\revise}{ The main issue is that the ViT needs a large amount of data to train due to a large number of parameters and lacks the inductive bias \cite{dosovitskiy2020image}. Existing FER datasets are much smaller compared to general image classification datasets (i.e. ImageNet \cite{deng2009ImageNet}), making it hard for newly proposed Transformer-based modules to converge well
and are easy to mistakenly focus on some occlusion or background areas. Many regularization \cite{srivastava2014dropout, ghiasi2018dropblock} and attention~\cite{Li2019OcclusionAware,8755326,wang2020region,xue2021TransFER,li2021MViTMask} methods have been proposed to address this issue.
}
The TransFER model generates an attention map and multiplies it with the feature maps to reduce the impact of noisy features. We investigated the TransFER model and
find that the model has learned to distinguish informative areas from noisy areas (as illustrated in Fig.~\ref{fig:la_attn}). However, the noisy features are still fed into the downstream models in TransFER. Hence, we raise a question: \textbf{Why do we still compute noisy features even though we have already known they are noises?}

Benefiting from the flexible design of the Transformer model, it could adopt any number of tokens as input without changing the model parameters. Inspired by this, we propose the APP module to discard the noisy features directly. As illustrated in the right part of Fig.~\ref{fig:la_attn}, the noisy features are now directly pooled (denoted as a small white grid) instead of multiplying with a small value.

\begin{figure}[t]
\begin{center}
\includegraphics[width=8.6cm]{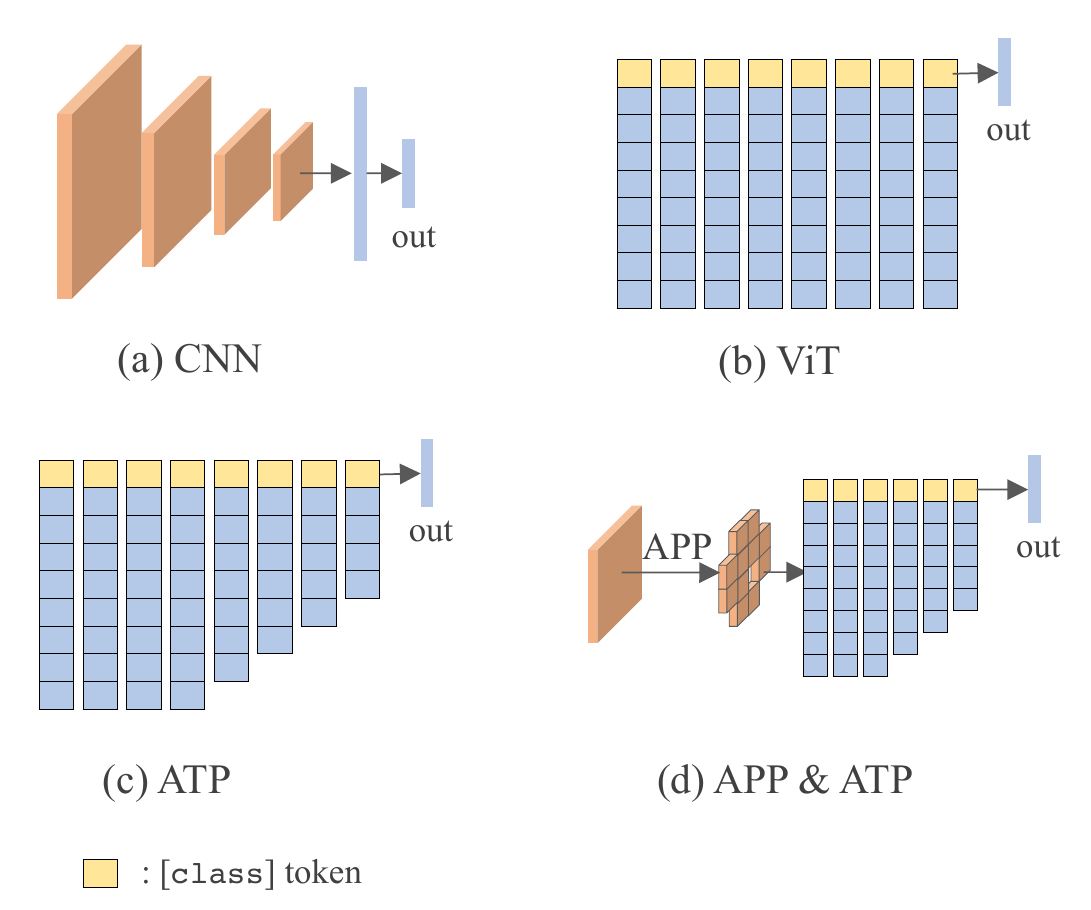}
\vspace{-10pt}
\end{center}
\caption{Illustration of the feature map size or token number of different architectures. (a) denotes the Max Pooling in CNN to quarter the feature map size. (b) denotes the vanilla ViT with a fixed token number. (c) denotes our ATP, which gradually reduces the number of tokens. (d) denotes our APViT, which consists of the APP and ATP to pool CNN features and ViT tokens simultaneously. }
\label{fig:diff}
\end{figure}

As for the Transformer block, it is built based on the attention mechanism, making it more intuitive to perform attentive pooling. Recently, CPVT \cite{Chu2021conditional} found that using the global average pooling (GAP) to replace the \texttt{[class]} token could produce an even better performance. DeepViT \cite{Zhou2021deepvit} further investigated this phenomenon, finding that the attention maps become similar after particular layers.
To reduce the redundancy in deep blocks, DeepViT \cite{Zhou2021deepvit} proposed a Re-attention method to increase the diversity of different layers, LV-ViT \cite{jiangAllTokensMatter2021} proposed a token labelling strategy to give a label to every token to supervise. Different from that, and inspired by the Max Pooling in CNNs (as illustrated in Fig.~\ref{fig:diff}~(a)) to reduce the size of feature maps and obtains an abstract and robust representation, we directly drop the less essential tokens and only remain a small number of tokens to embed the information.
Specifically, the ATP module utilizes the attention mechanism in the Transformer and directly discards the less important tokens in deep blocks. 
For a better understanding, the differences between these methods are illustrated in Fig.~\ref{fig:diff}. 

Fig.~\ref{fig:diff}~(b) demonstrates the vanilla ViT model where the patch token number remains constant. To interrupt the propagation of noisy features, the ATP gradually decreases the token number after a specific layer, as illustrated in Fig.~\ref{fig:diff}~(c). Compared with the Max Pooling in CNN models, which can only be pooled to a quarter at least, the ATP decreases the feature map size in a more controllable way.
Our proposed APViT model that combines APP and ATP is illustrated in Fig.~\ref{fig:diff}~(d).

Our contributions include the following:
\begin{enumerate}
\item An Attentive Patch Pooling (APP) module is proposed to select the most distinguishable local patches from the CNN feature maps,  which could prevent the noisy patches from passing through to the downstream modules to affect the recognition performance.
\item An Attentive Token Pooling (ATP) module is developed to utilize the attention mechanism of the Transformer model only to pay attention to the top-k relevant tokens to reduce the influence caused by noises and occlusions and save the computation time.
\item Experimental results on several challenging in-the-wild datasets demonstrate the advantages of our APViT method over the state of the arts. 

\end{enumerate}

\begin{figure*}[t]
\begin{center}
\includegraphics[width=0.9 \textwidth]{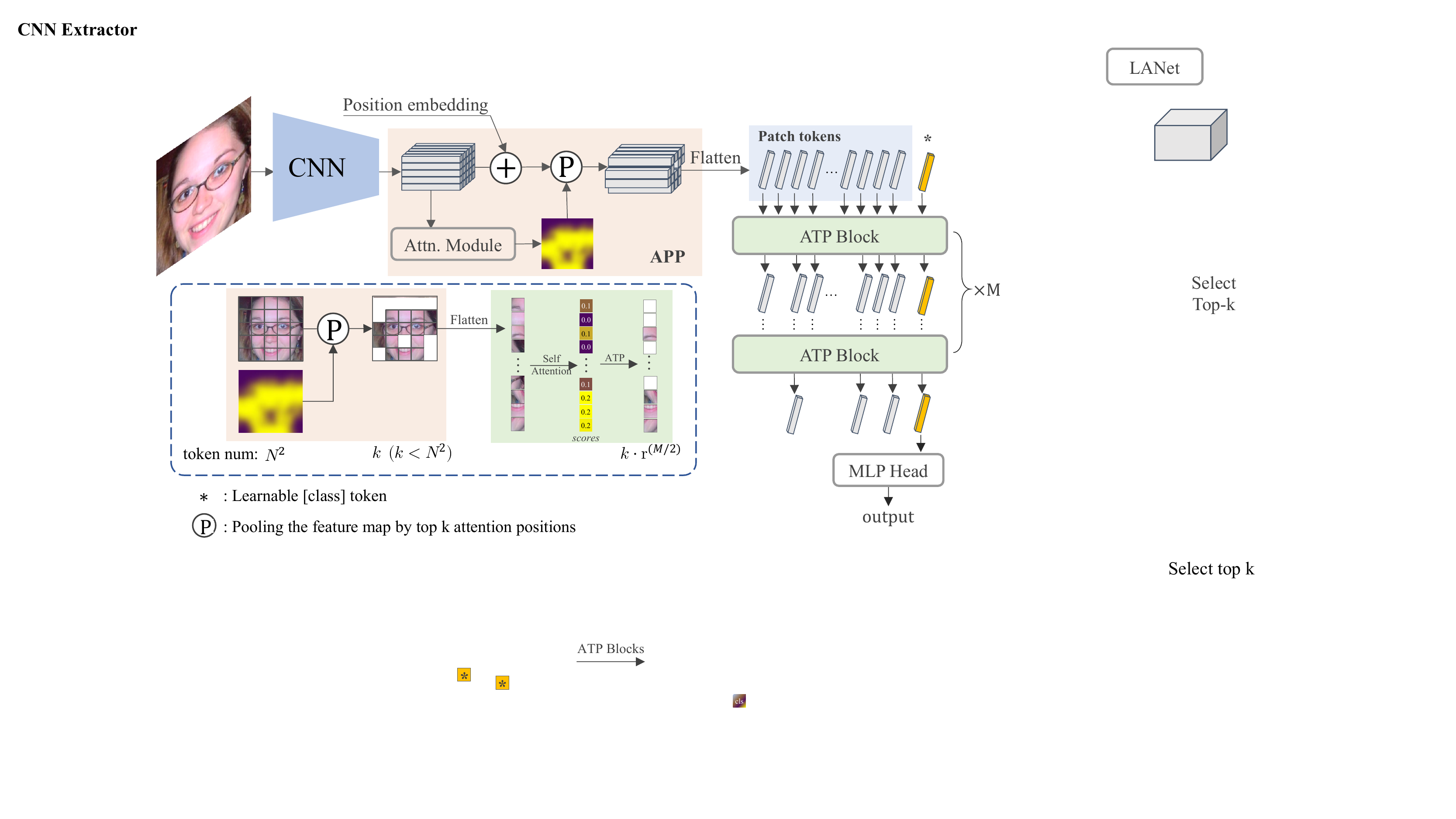}
\vspace{-10pt}
\end{center}
\caption{The overall architecture of our APViT model. Firstly, feature maps are extracted from the facial images by a CNN extractor (e.g. ResNet\cite{he2016deep}, MobileNet \cite{howard2017mobilenets}). Secondly, an attention map is calculated by the attention module ($attn\_f$), which guides the  APP module to select corresponding informative tokens. 
Thirdly, the selected tokens are flattened and fed into $M$ stacked ATP enhanced Transformer Blocks (denote as ATP Blocks) with an additional learnable \texttt{[class]} token attached. Similar to the APP, the ATP Blocks further select the most discriminative tokens with the multi-head self-attention mechanism. Thus, the patch information would be gathered to the \texttt{[class]} token as the number of kept patch tokens decreases gradually.
Finally, an MLP Head is attached to generate the final classification result from the \texttt{[class]} token. 
The left bottom blue dashed rectangle indicates the pooling process from the view of feature information flow.
The APP and ATP select the most distinguished patches and tokens, avoiding the noisy and occluded regions passing to the downstream calculation.
}
\label{fig:model}
\end{figure*}

\section{Related Work}

Facial expression recognition (FER) frameworks mainly consist of three processes: face detection and alignment, feature extraction, and classification. To reduce the impact of noisy background, a face detector (such as MTCNN~\cite{zhang2016joint} and RetinaFace~\cite{dengRetinaFace2020}) first detects faces from the image and aligns them to canonical alignments with corresponding detected landmarks. After that, a feature extractor is responsible for extracting discriminative features from the facial image. These features are then classified into expression categories. 
In the past decades, FER usually relied on well-designed hand-crafted features, which could be further grouped into texture-based and geometry-based features. For example, LBP \cite{shan2009facial}, Gabor~\cite{liu2002gabor}, HOG \cite{dalal2005histograms}, and SIFT \cite{ng2003sift} are proposed to utilize texture-based features. On the other hand, many geometry-based methods~\cite{happy2014automatic,ghimire2013geometric,saeed2014frame} take consider of features of some typical areas (such as eyes, noses, and mouths) based on landmarks. 
Most of these methods perform well on laboratory-collected databases. However, when facing in-the-wild facial images, such as various poses and occlusions, their performances are not as good as recently proposed learning-based methods~\cite{li2017reliable,cai2018IslandLoss,Li2019OcclusionAware,zeng2018facial,wang2020region,fanFacialExpressionRecognition2020}. With the help of public large-scale in-the-wild facial expression datasets, 
deep learning methods have achieved many breakthroughs.
Li \etal~\cite{li2017reliable} proposed the DLP-CNN method to extract discriminative features while Cai \etal~\cite{cai2018IslandLoss} proposed a island loss function to achieve the similar goal. Li \etal~\cite{Li2019OcclusionAware} utilized an attention mechanism to let the CNN network focus on the most discriminative un-occluded regions to overcome the occlusion problem.

Our proposed attentive pooling modules are mainly based on the patch attention mechanism and could affect tokens in the Transformer model. So we review the previous works in these two respects as follows:

\subsection{FER Based on Patch Attention}
\textcolor{\revise}{For high-resolution images, ATS~\cite{katharopoulos2019processing} samples locations of informative patches from an attention distribution based on a lower resolution image to save computation costs. NTS-Top~\cite{cordonnier2021differentiable} improves this attention sampling with a differentiable top-K patch selection module. STTS~\cite{wang2021efficient} selects tokens in both temporal and spatial dimensions to save computation in video tasks based on a scorer network and a differentiable top-K selection. For FER task, }
the facial expression is invoked by the movements of several facial muscles. Some researchers found that few patch areas contribute mostly to expression recognition \cite{duchenne1862mechanisme}. Many FER methods \cite{zhong2012LearningActiveFacial,happy2014automatic,xie2019deep} split the face image into patches and select discriminative regions with an attention mechanism.
Zhong~ \etal~\cite{zhong2012LearningActiveFacial} proposed a two-stage multi-task sparse learning framework to locate common patches shared by different expressions and specific patches useful for a particular expression. Happy~\etal~\cite{happy2014automatic} adopted the one-against-one classification method to determine salient facial patches. Nineteen patches around the eyes, nose, and mouth are pre-selected as candidate patches. A patch is selected if it can classify the two expressions accurately. 

Recent deep learning methods utilize attention mechanisms that learn an attention weight and multiply this weight with the features. Li~\etal~\cite{Li2019OcclusionAware} selected 24 patches based on facial landmarks and proposed the Patch-Gated Unit to perceive occluded patches.
Zhao~\etal~\cite{zhao2021LearningDeepGlobal} divided extracted feature maps into several local feature maps and employed a Convolution Block Attention Module \cite{woo2018cbam} to generate attention maps. The generated attention maps are multiplied by the original local feature maps to propagate gradients. \textcolor{\revise}{MViT~\cite{li2021MViTMask} utilizes two transformer modules where one generates a mask to filter out complex backgrounds and occlusion patches, and the other is the vanilla ViT model to perform the classification. 
However, background features in these methods still participate in the following calculation even though we know that they are less critical and may cause misfocus to be prejudicial to recognition.} Different from this, our Attentive Pooling modules directly discard these unimportant features, which could prevent the noisy information from passing to the following module and guide the model to ONLY focus on the most informative features. \textcolor{\revise}{
Another difference between our method and these methods is that we find the hand-designed mask generator performs better than the learning-based one. And benefitting this, our APViT could apply the origin top-K operation to replace the fancy differentiable ones.}

\subsection{Vision Transformers on FER}
The Transformer architecture \cite{vaswani2017attention} was first proposed for machine translation based on attention mechanisms.
And it is interesting to explore the Transformer to visual tasks \cite{carion2020DETR,zhu2020DeformableDETR}. Vision Transformer (ViT) \cite{dosovitskiy2020image} is the first work to directly apply the vanilla Transformer with few modifications for image classification. By pre-training on large datasets (e.g. ImageNet-1k, ImageNet-21k \cite{deng2009ImageNet}, and JFT \cite{sun2017JFT}) and fine-turning on the downstream target dataset, ViT has achieved excellent results compared to the CNN networks~\cite{touvron2020training, dascoliConViTImprovingVision2021,liuSwinTransformerHierarchical2021}. 

Some researchers also introduced ViTs to FER. Ma~\etal~ \cite{ma2021RobustFacial} proposed an attention selective fusion module to leverage feature maps which are generated by two-branch CNNs, and applied the vanilla ViT module to explore relationships between visual tokens.
Li~\etal~\cite{li2021MViTMask} proposed the MViT, utilizing two transformer modules where one generates a mask to filter out complex backgrounds and occlusion patches, and the other one is the vanilla ViT model to perform the classification.  A squeeze and excitation module \cite{aouayeb2021LearningVisionTransformer} was proposed before the MLP head in the ViT module to optimize the learning process on small facial expression datasets. TransFER \cite{xue2021TransFER} extracts abundant attention information with multi-branch local CNNs and the multi-head self-attention in the ViT and forces these attention features to be diverse by a multi-attention dropping module.

Unlike these methods, we proposed two attentive pooling methods without bringing extra training parameters to the ViT model, fully using the pre-trained weights. The pooling operation also forces the model only to pay attention to the most discriminative features, reducing the influence of occlusions and noises and saving the computation time.

\section{Proposed Method}

\subsection{Overview}

We propose two attentive pooling modules to gradually pool features in the convolutional neural network and vision transformer. As discussed above, our primary motivation is to select informative features and drop the features of background and occlusion with as few additional parameters as possible to overcome the over-fitting problem caused by limited data.

Fig.~\ref{fig:model} illustrates the framework of our proposed attentive pooling methods based on the hybrid ViT-small model \cite{dosovitskiy2020image}.
A facial image is first input into a CNN backbone to extract feature maps, then flattened and fed into multiple Transformer Blocks. 

Besides generating the patch tokens, the feature maps extracted from the CNN are encoded to an attention map by the attention generating method (denoted as $attn\_f$). 
The proposed APP module takes this attention map as input and selects the top-k patch tokens according to their corresponding attention values.

The selected patch tokens and the \texttt{[class]} token are jointly fed into the Transformer blocks~\cite{vaswani2017attention} which consist of alternating layers of multi-head self-attention (MSA) and MLP blocks to explore the relationship between different patches in a global scope. The ATP module is proposed for the Transformer Blocks to gradually decrease the token number, forcing the model to pay attention to discriminative tokens and eliminate the noisy tokens about background and occlusions.
After the Transformer encoder, the \texttt{[class]} token is used to generate the final predictions through a single fully connected layer.

\subsection{Attentive Patch Pooling}

Since the CNNs slide through the whole face, features from uninformative areas may introduce noises to the feature maps, which may be harmful to recognition. Our APP module is employed to locate informative areas and drop the uninformative ones to boost recognition performance.

The shape of the feature maps extracted from the CNN backbone is $[H, W, C]$ where $H$, $W$, and $C$ represent the feature maps' height, width, and channel number, respectively. 
The APP module takes these attention maps as input and selects the top-k most important patches from the total $H \times W$ patches according to the important weight of each position. 
To achieve this, we need to define a pooling criterion to guide where to pool and a pooling method to process pooling.

\textbf{Pooling Criterion.}
To determine which parts to pool, we need a pooling criterion to denote the importance of every pixel. Each pixel in the feature map denotes a small patch of the original facial input because the Max Pooling layers enlarge the reception field. So given the feature maps in $[H, W, C]$, an attention map with a size $[H, W, 1]$ is needed to denote the relevant weight. In other words, the pooling criterion $\mathcal{F}$ is responsible for reducing the channel dimension of the feature maps to one.

\begin{equation}
\mathcal{F} : R^{H \times W \times C} \xrightarrow[]{}  R^{H \times W \times 1}
\end{equation}

 There are a few methods that could complete this task. We group them into hand-designed generators and learning-based generators.

\begin{figure}[t]
\begin{center}
\includegraphics[width=7.5cm]{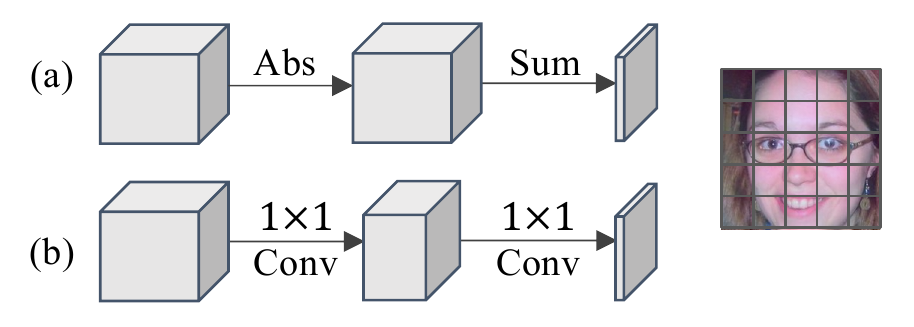}
\vspace{-10pt}
\end{center}
\caption{Illustration of the two kinds of pooling criteria. \textcolor{\revise}{(a) denotes the hand-designed generator, which calculates the sum of the absolute values of the feature maps. (b) denotes the learning-based generator. Taking the LANet \cite{Wang2020LANet} as an example, two $1 \times 1$ convolutional layers gradually decrease the channel dimension to 1.}}
\label{fig:CNNPooling}
\vspace{-10pt}
\end{figure}

\begin{enumerate}
 \item \textbf{Hand-designed generators} utilize mathematical operations to reduce the channel dimension. For simplicity, the sum of values ($F_{SUM}$), the sum of absolute values ($F_{ABS}$), and the max of values ($F_{MAX}$) \cite{Zagoruyko2017paying} are introduced here. 
As illustrated in Fig.~\ref{fig:CNNPooling} (a), the $F_{ABS}$ sums the absolute values of the feature maps along the channel dimension. The generated attention map puts high weight on locations where multiple channels respond with high activations. Similarly, the $F_{MAX}$ takes the maximal value among all neuron activations.
    
\item \textbf{Learning-based generators} utilize neural networks to learn the relationship between the feature maps of the attention map. The $1 \times 1 $ convolutional layer is widely used for this operation to reduce the channel dimension without changing the spatial resolution. The LANet \cite{Wang2020LANet} illustrated in Fig.~\ref{fig:CNNPooling} (b) is one of the examples of these generators. It combines two convolutional layers to gradually reduce the channel dimension. The sigmoid layer at the end of the original LANet is discarded here because only the relative relationship matters.
\end{enumerate}

\textbf{Pooling Method.}
After getting the attention map, we need a pooling method to drop the less relative patches and retain the most relative ones. A common method is to multiply the feature maps by the attention map, like in \cite{Wang2020LANet, xie2019deep}. However, this method fails to discard the noisy patches. Although locations with lower attention weights have smaller feature values, they are still computed in the incoming modules, which may bring noise and lead to misfocus.

The proposed APP module takes the top-k attention activation positions and selects the corresponding features. The detailed process is illustrated in Fig.~\ref{fig:model}.

\subsection{Attentive Token Pooling}

The APP module is efficient in pooling the uninformative and noise patches by the attention mechanism, but it lacks the global scope because of the small reception field of CNN. As illustrated in Fig.~\ref{fig:model}, the Transformer encoder enhanced by ATP (denoted as ATP Block) is utilized to explore the relationship among selected patches in a global scope. We start with a brief description of the Transformer Encoder.

\textbf{Transformer Encoder.}
The Transformer model is firstly proposed to process natural language processing (NLP) tasks and then utilized for image recognition \cite{dosovitskiy2020image}. It contains a stack of encoder blocks, with each block mainly consisting of Multi-head Self-Attention (MSA), Multi-Layer Perception (MLP), and Layer Normalization (LN).
The MSA module takes a series of tokens as input, and linearly transforms them into three parts, denoted as queries ($Q \in \mathbb{R}^{(n+1) \times d}$), keys ($K \in \mathbb{R}^{(n+1) \times d}$), and values ($V \in \mathbb{R}^{(n+1) \times d}$), where $n$ is the sequence length of the tokens, and $d$ is the embedding dimensions. The \texttt{[class]} token is concatenated to the input tokens, so the total length of the input is $n+1$. Every token is updated based on the correlation with each other. It can be considered a weighted sum calculated by the dot product of $Q$ and $K$. This process can be formulated as:

\begin{equation}
    Attention(Q, K, V) = softmax(\frac{QK^T}{\sqrt{d}})V \label{Equ:qkv}
\end{equation}

After the Transformer encoder, only the \texttt{[class]} token is converted by a single-layer MLP layer to generate the output score.

As we can see, the Transformer encoder does not change the length of the input sequence, nor do the MLP or the LN. It is different from the CNN models, where the Max Pooling is widely used to reduce the spatial resolution and increase the generalization ability. 
So, we propose the ATP directly drop the less important ones and only remain a small number of tokens to embed the information. For a better understanding, the differences between these methods are illustrated in Fig.~\ref{fig:diff}. 

\textbf{Pooling Criterion.}
Like the APP described above, the ATP also needs a pooling criterion to guide where to pool. However, unlike the CNN, the Transformer architecture is built based on the attention mechanism. The Transformer model is much more intuitive to find where to pay more attention without an additional attention-generating module.

In the self-attention module, every token's query will compute a relation with the key of other tokens by a dot product, as formulated in Eq.~(\ref{Equ:qkv}). The output of a  token is calculated by a weighted sum of the values of input tokens. Moreover, the weight is determined by the query and key. The higher the $QK^T$ value is, the more relevant between these two tokens.
As described above, the \texttt{[class]} token is utilized to represent the whole image.
The more relative between some patch tokens and the \texttt{[class]} token, the more important these patch tokens are to recognize the expression.
So we take the $QK^T$ of the \texttt{[class]} token as the pooling criterion to remove the less relative tokens. As there are multiple heads, the sum of all heads is utilized.

\textbf{Pooling Method.}
The pooling method in ATP is similar to APP. We only consider the top-k tokens and remove the rest based on the pooling criterion. It is worth noting that the pooling only acts on patch tokens, and the \texttt{[class]} token is always reserved. Because there are several stacked attention blocks in the Transformer encoder, we propose a keep rate $r$ to adjust the pooling process, which is defined as follows:

\begin{equation}
    keep\_num = \lfloor r * input\_token\_number \rfloor
\end{equation}
where $\lfloor x \rfloor$ denotes the floor operation. As discovered in DeepViT, the attention collapse issue only happens in deep blocks, so the ATP is only applied to the second half of Transforme blocks. We set $r$ in these blocks to the same value for convenience. Specifically, denote there are $M$ blocks. The output token number will be about $k \cdot r^{M/2}$, ignoring the influence of the \texttt{[class]} token.

\section{Experiments}

\subsection{Datasets}
We evaluate our proposed method on six in-the-wild FER datasets: RAF-DB~\cite{li2017reliable}, FERPlus~\cite{barsoum2016training},   AffectNet~\cite{mollahosseini2017affectnet}, SFEW~\cite{dhallStaticFacialExpression2011}, ExpW~\cite{Zhang2018ExpW}, and Aff-Wild2~\cite{kollias2021analysing}.

\textbf{RAF-DB} contains 29,672 real-world facial images collected by searching on Flickr and was labelled by 40 trained human workers. In our experiments, the basic annotation images are utilized, resulting in 12,271 images for training and 3,068 images for testing, with six basic expressions and neutral.

\textbf{FERPlus} is extended from FER2013 \cite{goodfellow2013challenges}, providing more accurate annotations relabeled by ten workers to eight expression categories (six basic expressions, neutral, and contempt). It contains 28,709 training images, 3,589 validation images, and 3,589 test images. 

\textbf{AffectNet} 
contains about 1M facial images collected by three major search engines, where about 420K images were manually annotated. Following the settings in \cite{li2021adaptively}, we only used about 280K training images and 3,500 validation images (500 images per category) with seven emotion categories for a fair comparison.

\textbf{SFEW 2.0} was created by extracting static frames from the Acted Facial Expressions in the Wild (AFEW) \cite{dhall2011acted}, a temporal facial expressions database. It was used as a benchmark in the third Emotion Recognition in the Wild Challenge (EmotiW2015), with 958 facial images for training and 436 372 images for validation and testing. Since the label on the test set is not public, the performance on the validation set is reported.

\textbf{ExpW} was collected by searching combined emotion-related keywords (such as "excited", "panic") and occupation nouns (such as "student", "layer") on the Google image search. Facial images are filtered by a facial detector and labelled into six basic expressions and neutral. There is no official training set split approach. For a fair comparison, we follow \cite{bishaySchiNetAutomaticEstimation2021} to split the dataset into 75\% for training, 10\% for validation, and 15\% for testing with random sampling. As a result, there are 68,845, 9,179, and 13,769 images for training, validation, and testing.

\textbf{Aff-Wild2}~\cite{kollias2021analysing,kollias2019expression} is a recently proposed, large-scale facial expression recognition database. 
It is an extension of the Aff-Wild database \cite{zafeiriou2017affwild} and by far is the largest public FER dataset in the wild. It contains 564 videos of around 2.8M frames and labels them with three representations, including Categorical Emotions (CE), Action Units (AU), and Valence Arousal (VA). Following \cite{thinhEmotionRecognition2021}, we adopt the training set of both CE and AU tasks for training and the validation set of the CE task for testing. For a fair comparison, we take the same head design of \cite{thinhEmotionRecognition2021}, but with a different loss weight. Specifically, we sum the expression classification loss and action unit loss with weights 10 and 1 to strengthen the classification performance.

\begin{table}[]
\centering
\caption{Evaluation (\%) of APP, ATP, \texttt{[class]} token (CLT), global average pooling (GAP) on RAF-DB and AffectNet, and  corresponding FLOPs compared with the baseline.}
\label{tab:ab:ab}
\begin{tabu}{cccccc}
\toprule
APP & ATP & Head & RAF-DB & AffectNet & FLOPs \\ \midrule
\rowfont{\color{\revise}}    &     &  GAP & 90.77  & 65.66  & 100\%  \\
    &   & CLT & 91.26  & 65.91 & 100\%  \\
     & $\checkmark$  & CLT  & 91.52  & 66.65 & 94\% \\
\rowfont{\color{\revise}} $\checkmark$ &   & GAP & 90.91 & 66.34 & 88\% \\
$\checkmark$ &   & CLT & 91.65  & 66.51 & 88\% \\
$\checkmark$ & $\checkmark$  & CLT & 91.98  & 66.94 & 83\% \\  \bottomrule

\end{tabu}
\end{table}

\begin{table}
\centering
\caption{Evaluation (\%) of APP and ATP on Aff-Wild2 and corresponding FLOPs compared with the baseline.}
\label{tab:ab:aff}
\begin{tabular}{cccc}
\toprule
\makecell[c]{APP} & \makecell[c]{ATP} &
Aff-Wild2  & FLOPs \\ \midrule
             &              & 85.02  & 100\%  \\
             & $\checkmark$ & 85.70  & 81\%  \\
$\checkmark$ &              & 85.39  & 61\%  \\
$\checkmark$ & $\checkmark$ & 86.03  & 54\%  \\  \bottomrule

\end{tabular}
\end{table}

\subsection{Implementation Details}
All images were aligned and cropped to $112\times112$ pixels by MTCNN~\cite{zhang2016joint} before being input into the model. The first three stages of IR50 \cite{deng2019arcface}, pre-trained on Ms-Celeb-1M
\footnote{The pre-trained weight is downloaded from \url{https://github.com/ZhaoJ9014/face.evoLVe.PyTorch\#Model-Zoo}.} \cite{guo2016ms} are used as the feature extractor. The ViT-small~\cite{dosovitskiy2020image} model with 8 Transformer blocks pre-trained on ImageNet\footnote{The pre-trained weight is downloaded from \url{https://github.com/rwightman/pytorch-image-models/}.} is used as Transformer Encoders. 
The keep number ($k$) of the APP is set to 160 for RAF-DB, AffectNet, and FERPlus; and 120 for SFEW. The keep rate ($r$) of ATP is set to 0.9 for these four databases. It is worth noting that, 
ExpW and Aff-Wild2 achieve the best performance with only $k = 80$ and  $r = 0.6$, indicating that our APViT works more efficiently on large-scale datasets.
It is worth noting that there are 105 tokens reserved before the classification head when $k = 160$ and  $r = 0.9$, and only 10 tokens reserved when $k = 80$ and  $r = 0.6$. Our APViT maintains the performance with about half computation costs.
Our model was trained with the SGD optimizer to minimize the cross-entropy loss. The momentum was set to 0.9, and the weight decay was set to 5e-4. The mini-batch size was set to 128, and the learning rate was set to 1e-3 for RAF-DB, FERPlus, SFEW, Aff-Wild2, and 5e-4 for AffectNet and ExpW with cosine decay \cite{Loshchilov2016SGDR}. We also found that gradient clipping of 10 based on the L2-norm is helpful, as described in \cite{dosovitskiy2020image}. 
During the training process, we follow the same augmentation methods as TransFER \cite{xue2021TransFER}.
At the test time, only the resize operation is reserved.
We train our model with one NVIDIA V100 GPU, 20 epochs for FERPlus, 40 epochs for RAF-DB and SFEW, 6k iterations for AffectNet, 9k iterations for ExpW, and 150k iterations for Aff-Wild2.
The overall accuracy is reported by default.

\subsection{Ablation Studies}

\begin{figure}[]
\begin{center}
\includegraphics[width=0.4\textwidth]{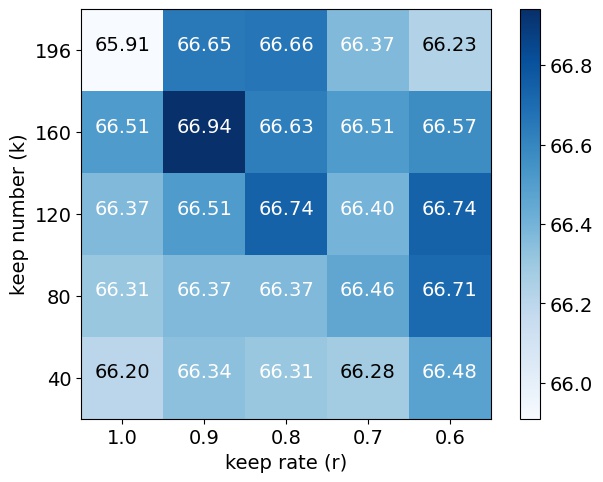}
\vspace{-10pt}
\end{center}
\caption{Evaluation (\%) of different keep numbers ($k$) and keep rates ($r$) on AffectNet. As the keep number and keep rate decade, the performance first increases and then decreases. From the top left to the bottom right, our AP modules reduce more than half of the computations, and the performance is robust without hard decreases, indicating the effectiveness of our AP modules.
}
\label{fig:ab:diff_k}
\end{figure}

\textbf{Effectiveness of the Proposed Modules.}
To evaluate the effect of the proposed modules, we designed the ablation study on RAF-DB and AffectNet to better understand the impact of the proposed APP and ATP. Our two pooling modules are designed to resolve the over-fitting problem when transferring the pre-trained Transformer model to small datasets and restrain the impact of noisy patches.

As is shown in Tab.~\ref{tab:ab:ab}, both modules increase the performance in FER compared with the baseline strategy. The APP and ATP contribute similarly when applied solely. The APP improves the performance by 0.39\% and 0.60\% in RAF-DB and AffectNet, respectively. The ATP brings a similar performance gain, 0.26\%, and 0.74\%, respectively. However, the APP takes effect more early, thus decreasing more FLOPs than ATP. The best improvement is obtained when combining two pooling methods. Specifically, the baseline is improved from 91.26\% to 91.98\% on RAF-DB and from 65.91\% to 66.94\% on AffectNet with only 83\% FLOPs\footnote{The number of FLOPs is calculated using the \textit{fvcore} toolkit provided by Facebook AI Research}. 
\textcolor{\revise}{
We also compared the \texttt{[class]} token (CLT) with global average pooling (GAP) with different strategies. Although GAP outperforms CLT in CVPT~\cite{Chu2021conditional}, it fails to achieve a better performance here. We suspect different pre-trained strategies cause this. The model should change as little as its pre-trained for downstream finetuning.
}
These experiments illustrate that the APP and ATP could help the model to only focus on more discriminative patches and reduce the over-fitting problem caused by the noisy patches.

\textbf{Impact of the Keep Number $k$ in APP and Keep Rate $r$ in ATP.}
The keep number ($k$) and keep rate ($r$) are two hyper-parameters of our proposed model to control how many patches to keep. To explore the impact of these two parameters, we study different values from 196 to 40 for $k$, and 1 to 0.6 for $r$ on AffectNet. When $k=196$ and $r=1$, it represents the baseline model with no pooling taking effect. The results are shown in Fig.~\ref{fig:ab:diff_k}, the best performance (66.94\%) could be obtained when $k=160$ and $r=0.9$. The smaller $k$ and $r$ are, the fewer patches are kept for recognition, which forces the model to focus on more discriminative patches and take less computation. Too less information, however, may make it harder to recognize, so the performance decreases when $k$ and $r$ are too small. It is worth noting that the minimal model ($k=40$ and $r=0.6$) here only has five tokens reserved but still outperforms the baseline model. Specifically, \textit{it boosts the performance from 65.91\% to 66.48\%, but with only 45.12\% FLOPs compared to the baseline model}.

\textcolor{\revise}{
\textbf{Evaluation of Pre-trained Weights.}
To evaluate the effects of pre-trained weights on CNN and ViT, we conduct experiments on RAF-DB and AffectNet. As illustrated in Tab.~\ref{tab:ab:pre-train}, the pre-trained weight is more critical for CNN, which boosts the performance from 81.85\% to 90.84\% on RAF-DB. This indicates that CNN pre-trained on facial recognition databases can extract helpful features for FER. In addition, pre-training on ViT has a negative influence when no CNN pre-trained weight is loaded but could help the model to achieve better performance with pre-trained CNN. 
}

\begin{table}
\centering
\color{\revise}\caption{Performances with or without pre-trained weights.}
\label{tab:ab:pre-train}
\color{\revise}\begin{tabular}{cccc}
\toprule
Pre-trained CNN & Pre-trained ViT & RAF-DB  & AffectNet \\ \midrule
             &              & 81.55  & 53.88 \\
$\checkmark$ &              & 90.84  & 65.34 \\
             & $\checkmark$ & 78.91  & 51.56 \\
$\checkmark$ & $\checkmark$ & 91.98  & 66.91 \\  \bottomrule
\end{tabular}
\end{table}

\begin{table}[t]
\centering
\color{\revise}\caption{Performances (\%) with gradually decayed and constant $k$ and $r$.}
\label{tab:ab:gradually}
\color{\revise}\begin{tabular}{ccc}
\toprule
Database  & Gradually decay  & Constant \\ \midrule
RAF-DB    &  91.46       & \textbf{91.98}  \\
FERPlus     & 90.19      & \textbf{90.86}  \\
AffectNet   & 66.08      & \textbf{66.91} \\ \bottomrule 
\end{tabular}
\end{table}

\textcolor{\revise}{
\textbf{Comparison of Gradually Decay $k$ and $r$.} As AP modules rely on accurate attention maps, however training from scratch cannot get accurate attention maps right from the start. Intuitively, not to drop patches in the early stage of training, gradually increasing the proportion of dropped patches as the network learns is a more appropriate training strategy. To evaluate the impact of this strategy with the constant one, we designed experiments with gradual decay of the $k$ and $r$ on RAF-DB, FERPlus, and AffectNet. Specifically, the $k$ and $r$ are initially set to 196 and 1.0 at the beginning and linearly decayed to the database's corresponding values.
Surprisingly, as we can see from Tab.~\ref{tab:ab:gradually}, the performances with gradually decayed $k$ and $r$ are slightly inferior to the constant ones. This may be because inaccurate attention maps perform like random drop out at the beginning of the training period, pushing the model to find the informative features and preventing the model from over-fitting to noisy backgrounds. However, the gradual decay strategy lacks this benefit.
}

\textbf{Comparison of Different Attention Generate Methods.}
As discussed in the Method section, there are several different methods to generate attention maps. We run an ablation study to specify the different impacts of these methods on RAF-DB and AffectNet. As illustrated in Tab.~\ref{tab:ab:diff_method}, when $F_{SUM}$ is adopted as the method in ATP, $F_{LANet}$ (adopted by TransFER \cite{xue2021TransFER}) performs the worst (91.23\% and 66.28\%) and even worse than the baseline (91.26\% ) on RAF-DB. This may be because the LANet needs to multiply the attention map with the features to backward the gradient. 
The multiplication operation changes the magnitude of features and makes the downstream network harder to adapt. Hand-designed methods do not have this problem and achieve better performance. The best performance (91.98\% and 66.94\%) is achieved by $F_{ABS}$, which indicates that \textit{the negative values in CNN features also contain rich semantic information}.

We also compare three hand-designed methods in ATP. The $F_{ABS}$ here does not achieve comparable performance as in APP did and is much lower than $F_{SUM}$. It is because the dot product attention mechanism has a strong mathematical meaning: small values (including negative ones) indicate a poor relationship. While taking the absolute value as attention value breaks this mechanism. The $F_{MAX}$ takes the max value among multiple heads, but certain heads may dominate the result, which reduces the discriminative ability.

\begin{table}
\centering
\caption{Evaluation (\%) of different pooling methods on RAF-DB, the method action on APP and ATP are denoted as APP method and ATP method, respectively.}
\label{tab:ab:diff_method}
\begin{tabular}{cccc}
\toprule
APP Method  & ATP Method            & RAF     & AffectNet \\ \midrule
LANet       & \multirow{4}{*}{SUM} & 91.23  & 66.28     \\
MAX         &                     & 91.36  & 66.51   \\
SUM         &                     & 91.72  & 66.66     \\
ABS         &                      & \textbf{91.98} & \textbf{66.94} \\ \hline
\multirow{2}{*}{ABS} & ABS  & 91.33  & 66.34  \\
                      & MAX       & 91.39  & 66.54 \\ \bottomrule 
\end{tabular}
\end{table}

\begin{table}[t]
\centering
\color{\revise}\caption{Performances with “soft” and “hard” pooling approaches.}
\label{tab:ab:soft}
\color{\revise}\begin{tabular}{ccc}
\toprule
Database  & Soft  & Hard (Ours) \\ \midrule
RAF-DB    &  91.07       & \textbf{91.98}  \\
FERPlus     & 90.55      & \textbf{90.86}  \\
AffectNet   & 65.80      & \textbf{66.91} \\ \bottomrule 
\end{tabular}
\end{table}

\begin{table}
\centering
\caption{Performance comparison (\%) with SOTA methods on RAF-DB and AffectNet. ``*" indicates Transformer-based models.}
\label{tab:sota:rafaffect}
\begin{tabular}{ccc}
\hline
Method                               & RAF-DB & AffectNet \\ \hline
DLP-CNN \cite{li2017reliable}        & 80.89  & 54.47     \\
gACNN \cite{Li2019OcclusionAware}    & 85.07  & 58.78     \\
IPA2LT \cite{zeng2018facial}         & 86.77  & 55.71     \\
LDL-ALSG \cite{chen2020LabelDistribution} &85.53&59.35    \\
RAN \cite{wang2020region}            & 86.90  & 59.50     \\
CovPool \cite{acharya2018covariance} & 87.00  & -         \\
SCN \cite{wang2020suppressing}       & 87.03  & 60.23     \\
DLN \cite{zhang2021LearningFacial}   & 86.40   & 63.70    \\
DACL \cite{farzaneh2021facial}       & 87.78  & 65.20     \\
KTN \cite{li2021adaptively}          & 88.07  & 63.97     \\ 
EfficientFace \cite{zhao2021robustlight}&88.36& 59.89     \\ 
PAT-Res101 + attr \cite{caiProbabilisticAttributeTree2022}&88.43& - \\
FDRL \cite{ruan2021feature}           & 89.47   &  -      \\ 
PT \cite{jiangBoostingFacialExpression2021} & 89.57 & 58.54  \\ \hline
VTFF* \cite{ma2021RobustFacial}        & 88.14  & 61.85  \\
MViT* \cite{li2021MViTMask}           & 88.62  & 64.57   \\
TransFER* \cite{xue2021TransFER}      & 90.91 & 66.23  \\ \hline
\textbf{  APViT* (Ours)} & \textbf{91.98} & \textbf{66.91} \\ \hline
\end{tabular}
\end{table}

\begin{table*}
\centering
\caption{Per-class performance comparison (\%) with the state-of-the-art methods on RAF-DB. ``*" indicates Transformer-based models.}
\label{tab:sota:raf_avg}
\begin{tabu}{c|ccccccc|c}
\toprule
Method        & Anger & Disgust & Fear  & Sadness & Happiness & Surprise & Neutral & Average \\ \midrule
VGG~\cite{li2021adaptively}  & 53.09 & 33.75   & 14.87 & 77.41   & 92.41     & 67.48    & 82.35   & 60.19   \\
ResNet~\cite{li2021adaptively} & 62.58 & 18.13   & 22.97 & 59.21   & 90.04     & 72.04    & 82.50   & 58.35   \\ \midrule
CL-CNN~\cite{wen2016discriminative} & 71.61 & 14.38   & 9.46  & 83.26   & 94.09     & 86.63    & 83.68   & 63.30   \\
DLP-CNN~\cite{li2017reliable} & 71.60 & 52.15   & 62.16 & 80.13   & 92.83     & 81.16    & 80.29   & 74.20   \\
IL-CNN~\cite{cai2018IslandLoss} & 70.99 & 45.63   & 32.43 & 76.78   & 91.48     & 78.42    & 75.74   & 67.35   \\
Boosting-POOF~\cite{liu2017boosting} & 73 & 57    & 64  & 74   & 89      & 80     & 76    & 73.19   \\
STSN~\cite{li2021adaptively} & 82.72 & 66.25   & 59.46 & 87.87   & 94.35     & 86.63    & 85.00   & 80.32   \\
KTN~\cite{li2021adaptively} & 81.48 & 65.62   & 68.92 & 87.24   & 94.60     & 83.28    & 88.53   & 81.38   \\ 
PT~\cite{jiangBoostingFacialExpression2021} & 81   & 55   & 54   & 87   & 96   & 87   & 92   & 78.86 \\
PAT-Res101 + attr \cite{caiProbabilisticAttributeTree2022} & 77.16 & 60.00 & 67.57 & 86.61 & 95.53 & 89.67 & 88.38 & 80.70 \\ \midrule
MViT*~\cite{li2021MViTMask} & 78.40 & 63.75   & 60.81 & 87.45   & 95.61     & 87.54    & 89.12   & 80.38   \\
VTFF*~\cite{ma2021RobustFacial} & 85.80 & 68.12   & 64.86 & 87.24   & 94.09     & 85.41    & 87.50   & 81.20   \\ 
\rowfont{\color{\revise}}TransFER*~\cite{xue2021TransFER} & \textbf{88.89} & \textbf{79.37}   & 68.92 & \textbf{88.70}   & 95.95     & 89.06    & 90.15   & 85.86   \\ \midrule
\textbf{APViT* (Ours)} & 86.42 & 73.75 & \textbf{72.97} & \textbf{88.70} & \textbf{97.30} & \textbf{93.31} & \textbf{92.06} & \textbf{86.36}  \\ \bottomrule
\end{tabu}
\end{table*}

\begin{figure*}
\begin{center}
\includegraphics[width=0.95 \textwidth]{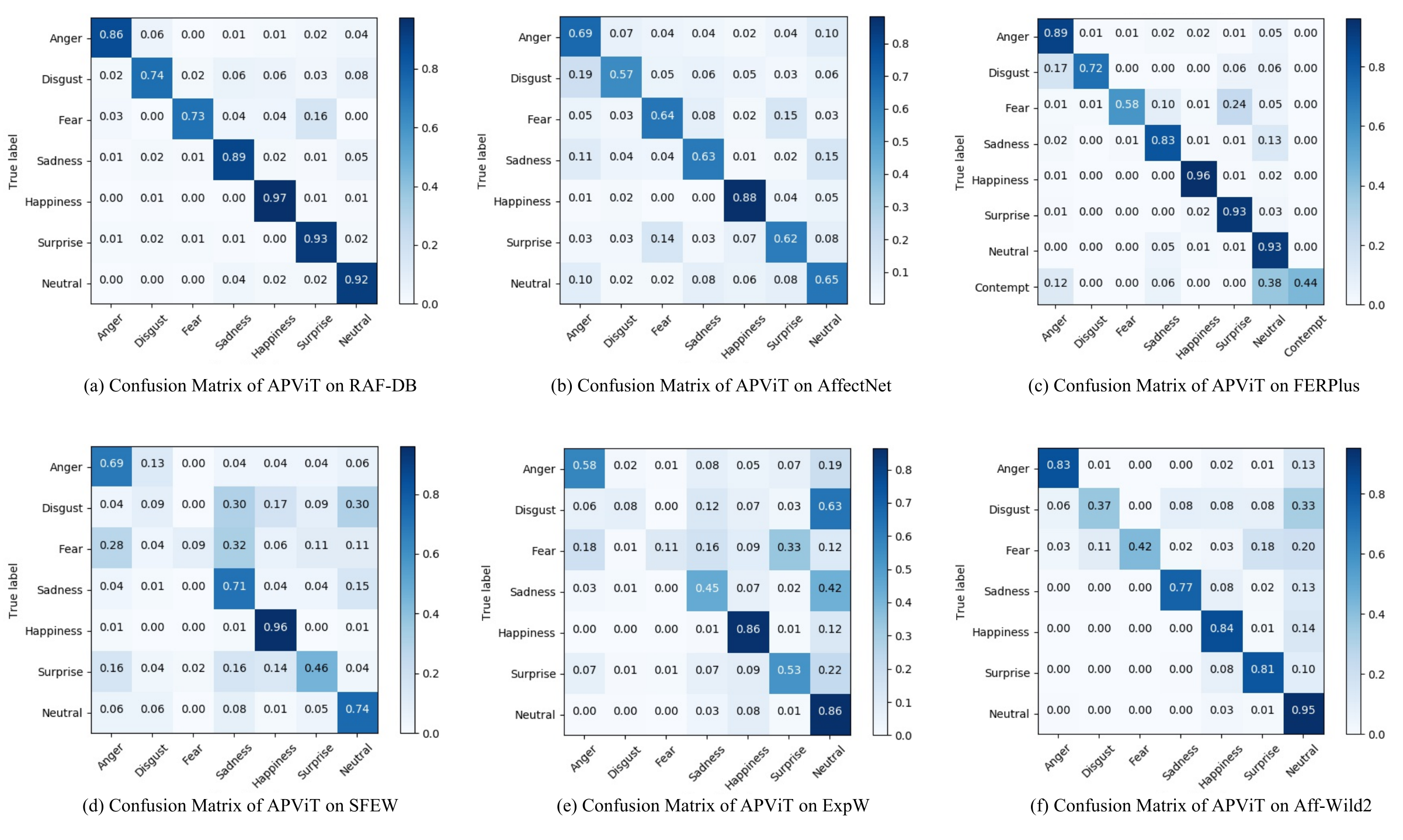}
\vspace{-10pt}
\end{center}
\caption{Confusion matrices of our APViT model on RAF-DB, AffectNet, FERPlus, ExpW, SFEW, and Aff-Wild2 databases. 
}
\label{fig:cm}
\end{figure*}

\textcolor{\revise}{
\textbf{Comparison of "Soft" Pooling with Our "Hard" Pooling.}
In most deep-learning-based models, like the LANet in the TransFER model, multiplying the mask with the features (denoted as "soft" pooling) is the only choice to propagate gradients to the mask-generating branch. We overcome this restriction by utilizing the hand-designed attention generator and found that its performance is good enough and could reduce the computation.  
To compare these two approaches, we conduct experiments on three databases, as illustrated in Tab.~\ref{tab:ab:soft}.
As we can see, the "soft" pooling approach performs much worse than the "hard" one, especially for the noisy database (AffectNet). The conclusion is the same as the LANet, which has been illustrated in Tab.~\ref{tab:ab:diff_method}. This indicates that our proposed "hard" pooling is more efficient in helping the model to find the informative features.
} 

\subsection{Comparison with the State-of-the-Art Methods}
In this section, we compare our proposed method with several state-of-the-art (SOTA) methods on six major in-the-wild datasets in Tab.~\ref{tab:sota:rafaffect}~-~\ref{tab:sota:aff}. We also illustrate the confusion matrices of these databases on Fig.~\ref{fig:cm}.

\textbf{Results on RAF-DB.} The comparison with other progressive methods on RAF-DB is illustrated in Tab.~\ref{tab:sota:rafaffect}. TFE~\cite{li2021FacialExpression} adopts joint learning with identity and emotion information. DLP-CNN \cite{li2017reliable}, DACL \cite{farzaneh2021facial}, and KTN \cite{li2021adaptively} utilize new proposed loss functions to enhance the discriminative power of deep features; IPA2LT \cite{zeng2018facial}, LDL-ALSG \cite{chen2020LabelDistribution}, and SCN \cite{wang2020suppressing} tend to increase the performance by solving the uncertain problem in the FER task. VTFF~\cite{ma2021RobustFacial}, MViT~\cite{li2021MViTMask} and TransFER~\cite{xue2021TransFER} use Transformer to obtain global-scoop attentions. Among that, TransFER obtained a big performance gain with 90.91\%. Based on their architecture, our proposed APViT further improves the recognition performance with the attentive pooling modules by restraining the noisy influence. We also find that realigning the facial images in the official RAF-DB database could further improve the performance. Our model outperforms the state-of-the-art models with 91.98\%.

The per-class results and the average accuracies on RAF-DB are illustrated in Tab.~\ref{tab:sota:raf_avg}. As we can see, our APViT outperforms other methods in most classes. Happiness has the highest accuracy with 97.30\%, and fear has the lowest accuracy with 72.97\% (but still outperforms other methods). The RAF-DB database is collected in the wild and is class-imbalanced because negative expressions (such as anger, disgust, fear, and sadness) are rare and challenging to collect in daily life. As is illustrated in Fig.~\ref{fig:cm}~(a), thanks to the reliable label annotation, our method performs well in negative expressions (the left top part of the confusion matrix). We believe this is because the attentive pooling module forces the model to find and focus on the most class-relative patch features.

\begin{table}[t]
\centering
\caption{Performance comparison (\%) with SOTA methods on FERPlus. ``*" indicates Transformer-based models.}
\label{tab:sota:ferplus}
\begin{tabular}{ccc}
\toprule
Method                               & FERPlus   \\ \hline
TFE-joint learning \cite{li2021FacialExpression}  & 84.29     \\
PLD \cite{barsoum2016training}       & 85.10     \\
RAN \cite{wang2020region}            & 88.55     \\
SeNet50 \cite{albanie2018emotion}    & 88.80     \\
RAN-VGG16 \cite{wang2020region}      & 89.16     \\
SCN \cite{wang2020suppressing}       & 89.35     \\
PT \cite{jiangBoostingFacialExpression2021} & 86.60 \\
KTN \cite{li2021adaptively}          & 90.49     \\ \hline
VTFF* \cite{ma2021RobustFacial}        & 88.81     \\
MViT* \cite{li2021MViTMask}           & 89.22     \\
TransFER* \cite{xue2021TransFER}      & 90.83  \\ \hline
\textbf{ APViT* (Ours)}     & \textbf{90.86}  \\ \bottomrule
\end{tabular}
\end{table}

\textbf{Results on AffectNet.} Since the testing set of AffectNet is not public. We evaluate and compare the performance of the validation set. As the training set of the AffectNet is imbalanced and the validation set is balanced, we employ a heavy over-sampling strategy as RAN~\cite{wang2020region}, SCN~\cite{wang2020suppressing}, MA-Net~\cite{zhao2021LearningDeepGlobal}, and TransFER does. For a fair comparison, the 7-categories results are illustrated in the third column of  Tab.~\ref{tab:sota:rafaffect}. As we can see, we obtain 66.91\% FER accuracy, which outperforms all other state-of-the-art methods. The confusion matrix, illustrated in Fig.~\ref{fig:cm}~(b), demonstrates that our model has a similar recognition capacity (57\% to 69\%) in all seven expression categories, while happiness has the highest accuracy with 88\%.

\textbf{Results on FERPlus.} Different from other databases, FERPlus has eight expression categories with an additional \textit{contempt}. As illustrated in Tab.~\ref{tab:sota:ferplus}, our method performs best with an accuracy of 90.86\%. Compared with TransFER, we maintain the performance with 83\% computations. The confusion matrix of FERPlus is illustrated in Fig.~\ref{fig:cm}~(c). As we can see, contempt performs worst and is hard to distinguish from neutral. This is because there are only 0.47\% contempt facial images versus about 29\% neutral images in the training set, making model predictions biased to neutral as illustrated in the right bottom of Fig.~\ref{fig:cm}~(c).

\begin{table}
\centering
\caption{Performance comparison (\%) with SOTA methods on SFEW.}
\label{tab:sota:sfew}
\begin{tabular}{ccc}
\toprule
Method                                    & SFEW 2.0 (Val) \\ \hline
DLP-CNN \cite{li2017reliable}             & 51.05 \\ 
Kim's CNN \cite{Kim2015Hierarchical}    &   52.5  \\
LBAN-IL \cite{Li2021LBAN}               & 55.28 \\
RAN(VGG16+ResNet18) \cite{wang2020region}  & 56.40  \\
APM~\cite{li2019pooling}                  & 57.57 \\
DAN \cite{wenDistractYourAttention2021}  & 57.88 \\
AHA \cite{WengAttentiveHybrid2021}      & 58.89 \\
MA-Net \cite{zhao2021LearningDeepGlobal}   & 59.40  \\
PAT-Res101 \cite{caiProbabilisticAttributeTree2022}  & 53.90 \\ 
PAT-Res101 + attr \cite{caiProbabilisticAttributeTree2022} & 57.57 \\ \hline
\textbf{ APViT (Ours)}     & \textbf{61.92}  \\ 
\bottomrule
\end{tabular}
\end{table}

\textbf{Results on SFEW.} We compare our APViT with several state-of-the-art methods on SFEW in Tab.~\ref{tab:sota:sfew}. LBAN-IT~\cite{Li2021LBAN} proposed a local binary attention network and an islets loss to discover local changes in the face. Since the training samples of SFEW are limited (less than 1,000 images for training), MA-Net~\cite{zhao2021LearningDeepGlobal} pre-trained the model in FER2013, while APM~\cite{li2019pooling} pre-trained the model in RAF-DB. With fine-tuning, they achieved performance with  59.40\% and 57.57\%. PAT-Res101~\cite{caiProbabilisticAttributeTree2022} extend the training dataset with RAF-DB to have 50\% RAF-DB samples in every mini-batch. PAT-Res101+attr~\cite{caiProbabilisticAttributeTree2022} further improved the performance by introduce attribute-annotated samples to 57.57\%. However, we only use the pre-trained weights on the RAF-DB database and outperform these methods without additional data with an accuracy of 61.92\%. As we can see from the confusion matrix, which is illustrated in Fig.~\ref{fig:cm}~(d), the accuracy of disgust and fear are both 9\% only. Since the distribution of SFEW is extremely imbalanced, there are only 54 disgust samples and 81 fear samples in the training set. The poor performance in these two categories is reasonable and acceptable. 

\begin{table}
\centering
\caption{Performance comparison (\%) with SOTA methods on ExpW.}
\label{tab:sota:ExpW}
\begin{tabular}{ccc}
\toprule
Method                                   & ExpW \\ \hline
HOG + SVM \cite{Zhang2018ExpW}           & 60.66 \\
Baseline DCN \cite{Zhang2018ExpW}               & 65.06 \\
DCN + AP \cite{Zhang2018ExpW}                   & 70.06 \\
Lian \etal \cite{lianExpressionAnalysisBased2020a}&71.90 \\ 
Pham \etal \cite{phamFacialActionUnits2019}              & 72.84 \\
PAT-Res101 \cite{caiProbabilisticAttributeTree2022}     & 69.29 \\
PAT-Res101 + attr \cite{caiProbabilisticAttributeTree2022} & 72.93 \\ 
SchiNet \cite{bishaySchiNetAutomaticEstimation2021}     & 73.10 \\ \hline
\textbf{ APViT (Ours)}     & \textbf{73.48}  \\ \bottomrule
\end{tabular}
\end{table}

\begin{table}
\centering
\caption{Performance comparison (\%) with SOTA methods on Aff-Wild2. }
\label{tab:sota:aff}
\begin{tabular}{ccc}
\toprule
Method                               & Aff-Wild2 (val) &  Rank    \\ \hline
Zhang \etal \cite{zhangPriorAidedStreaming2021a} & 85.6  & 1   \\
Jin \etal \cite{jinMTMSNMultiTaskMultiModal2021}& 63.0  & 2    \\
Thinh \etal \cite{thinhEmotionRecognition2021}   & 82.6   & 3   \\
Wang \etal \cite{wang2021multitask}      & 50.1    & 4   \\
Deng, Wu, and Shi \cite{dengIterativeDistillationBetter2021}    & 57.79    & 5 \\ \hline
\textbf{ APViT (Ours)}     & \textbf{86.03}  & -  \\ \bottomrule
\end{tabular}
\end{table}

\textbf{Results on ExpW.} Same as SFEW, due to the severe imbalanced problem, the performance in disgust and fear are poor, and predictions are biased to neutral as illustrated in Tab.~\ref{fig:cm}~(e). The comparison of our APViT with other state-of-the-art methods on ExpW is illustrated in Tab.~\ref{tab:sota:ExpW}. Lian~\etal~\cite{lianExpressionAnalysisBased2020a} divided the face into six areas to explore the authenticity of \textless emotion, region\textgreater pairs. They only chose the facial images with confident scores greater than 60 in the database and achieved 71.90\% accuracy. Pham~\etal~\cite{phamFacialActionUnits2019} used action unit features to find the images with a similar emotion for oversampling. SchiNet~\cite{bishaySchiNetAutomaticEstimation2021} combined ExpW with other databases (CEW~\cite{song2014eyes}, CelebA~\cite{liu2015deep}, and EmotionNet~\cite{fabian2016emotionet}) and achieved a promising accuracy, 73.10\%. With only training on the target database, our APViT outperforms these methods with 73.48\% accuracy. Moreover, benefiting from the large-scale database, our APViT achieves the best performance with only about half of (54.07\%) computations.

\textbf{Results on Aff-Wild2.}
We compare our proposed APViT with other state-of-the-art methods in the second ABAW2 Competition \cite{kollias2021analysing}, as illustrated in Tab.~\ref{tab:sota:aff}. As the test set is private, we only report the classification accuracy on the validation set. The rank column indicates their leader board rank on the test set. \cite{zhangPriorAidedStreaming2021, wang2021multitask, dengIterativeDistillationBetter2021} utilized all three annotations, including CE, AU, and VA, achieving 85.6\%, 50.1\%, and 57.79\% on the validation set, respectively. \cite{thinhEmotionRecognition2021} utilized only CE and AU and achieved a promising performance of 82.6\%. \cite{jinMultimodalMultitaskLearning2021} took the audio information as another modal and achieved 63.0\%. Following \cite{thinhEmotionRecognition2021}, we only take the CE and AU as input. With the help of our proposed two attentive pooling modules to reduce the influence of the noisy features, our APViT outperforms these state-of-the-art methods with an accuracy of 86.03\% on the validation set. And as illustrated in Fig.~\ref{fig:cm}~(f), the accuracy of disgust and fear is higher than that of SFEW and ExpW due to more training samples in these categories.

\begin{table}
\centering
\caption{Comparison of accuracy and computation cost against state-of-the-art methods on RAF-DB. (160, 0.9) indicates $k=160$, $r=0.9$ in APViT.}
\label{tab:ab:flops}
\begin{tabular}{lcc}
\toprule
Method          & GFLOPs & RAF-DB \\ \midrule
TransFER~\cite{xue2021TransFER}    & 15.30  & 90.91  \\
APViT (160, 0.9) & 12.67  & \textbf{91.98}  \\ \midrule
PAT-CNN~\cite{caiProbabilisticAttributeTree2022}    & 7.64   & 88.43  \\
KTN~\cite{li2021adaptively} & 7.60   & 88.07  \\
VTFF~\cite{ma2021RobustFacial}     & 6.08   & 88.14  \\
MViT~\cite{li2021MViTMask}  & 5.95   & 88.62  \\
APViT (10, 0.6)  & \textbf{5.89}     & 90.67 \\ \bottomrule
\end{tabular}
\end{table}

\begin{table}[t]
\centering
\caption{Performance comparison (\%) on lighter Transformer models on RAF-DB.}
\label{tab:ab:lighter}
\begin{tabular}{lccc}
\toprule
Model                 & Input     & GFLOPs & RAF-DB \\ \midrule
Hybrid-DeiT-Tiny      & 112$^2$   & 6.72  & 90.91    \\
Hybrid-DeiT-Tiny + AP & 112$^2$   & 6.10  & 91.26    \\
DeiT-Tiny             & 224$^2$   & 1.26  & 85.72    \\
DeiT-Tiny + AP        & 224$^2$   & 0.96  & 86.70 \\ 
\bottomrule
\end{tabular}
\end{table}

\subsection{Further Analysis}
\textbf{Accuracy and computation comparison with state-of-the-art methods.} To better demonstrate the effectiveness of our proposed attentive pooling modules, we illustrate the GFLOPs and accuracy on RAF-DB in Tab.~\ref{tab:ab:flops}. PAT-CNN~\cite{caiProbabilisticAttributeTree2022} and KTN~\cite{li2021adaptively} are CNN-based methods; TransFER~\cite{xue2021TransFER}, VTFF~\cite{ma2021RobustFacial}, and MViT~\cite{li2021MViTMask} are based on the Transformer. Our APViT (160, 0.9) achieves the best performance (91.98\%) with 82.81\% FlOPs compared with TransFER. And when we pool more patches and tokens to keep comparable FLOPs with other methods, our APViT (10, 0.6) outperforms them with 2.05\% higher accuracy. This indicate that our attentive pooling module drops background and noises and forces the model to only focus on discriminative information.

\textbf{Experiments on Lighter Transformer models.}
Tab.~\ref{tab:ab:lighter} shows the results on RAF-DB by exploring our two AP modules to two versions of lighter transformer models. The Hybrid-DeiT-Tiny model replaces the ViT-Small in APViT with DeiT-Tiny \cite{touvron2020training}. Our AP reports 0.35\% and 0.98\% performance improvements and 9.18\%, 23.68\% computation reduction on Hybrid-DeiT-Tiny and DeiT-Tiny, respectively. The $k$ and $r$ for Hybrid-DeiT are 120 and 0.9, respectively; and the $r$ for DeiT-Tiny is 0.8. Without the APP module, the ATP in DeiT-Tiny needs to pool more noisy features to get the best performance.

\textcolor{\revise}{\textbf{Experiments on CIFAR-10.} Most FER methods might not be suitable for general image classification tasks. This is mainly because face images are cropped and aligned before input into the FER model, making faces' position fixed and unified. While for general image classification tasks, foreground objects may occur at any place and scale. To investigate the generalization ability of our APViT model, we conduct experiments on CIFAR-10~\cite{krizhevsky2009learning}. As shown in Tab.~\ref{tab:ab:cifar}, our APViT maintains the performance with only 54.07\% computation costs and could slightly increase the performance from 96.02\% to 96.23\%, indicating the good generalization of the proposed method.
}

\begin{table}[t]
\centering
\color{\revise}\caption{Performance and GFLOPs comparison (\%) on CIFAR-10.}
\label{tab:ab:cifar}
\color{\revise}\begin{tabular}{cccc}
\toprule
$k$                 & $r$     & CIFAR-10 & FLOPs  \\ \midrule
\multicolumn{2}{c}{baseline}    & 96.02  & 100\%   \\
120      & 0.9   & 96.23  & 70.85\%    \\
80       & 0.6   & 96.11  & 54.07\%  \\ \bottomrule
\end{tabular}
\end{table}

\begin{figure}[t]
\begin{center}
\includegraphics[width=0.45\textwidth]{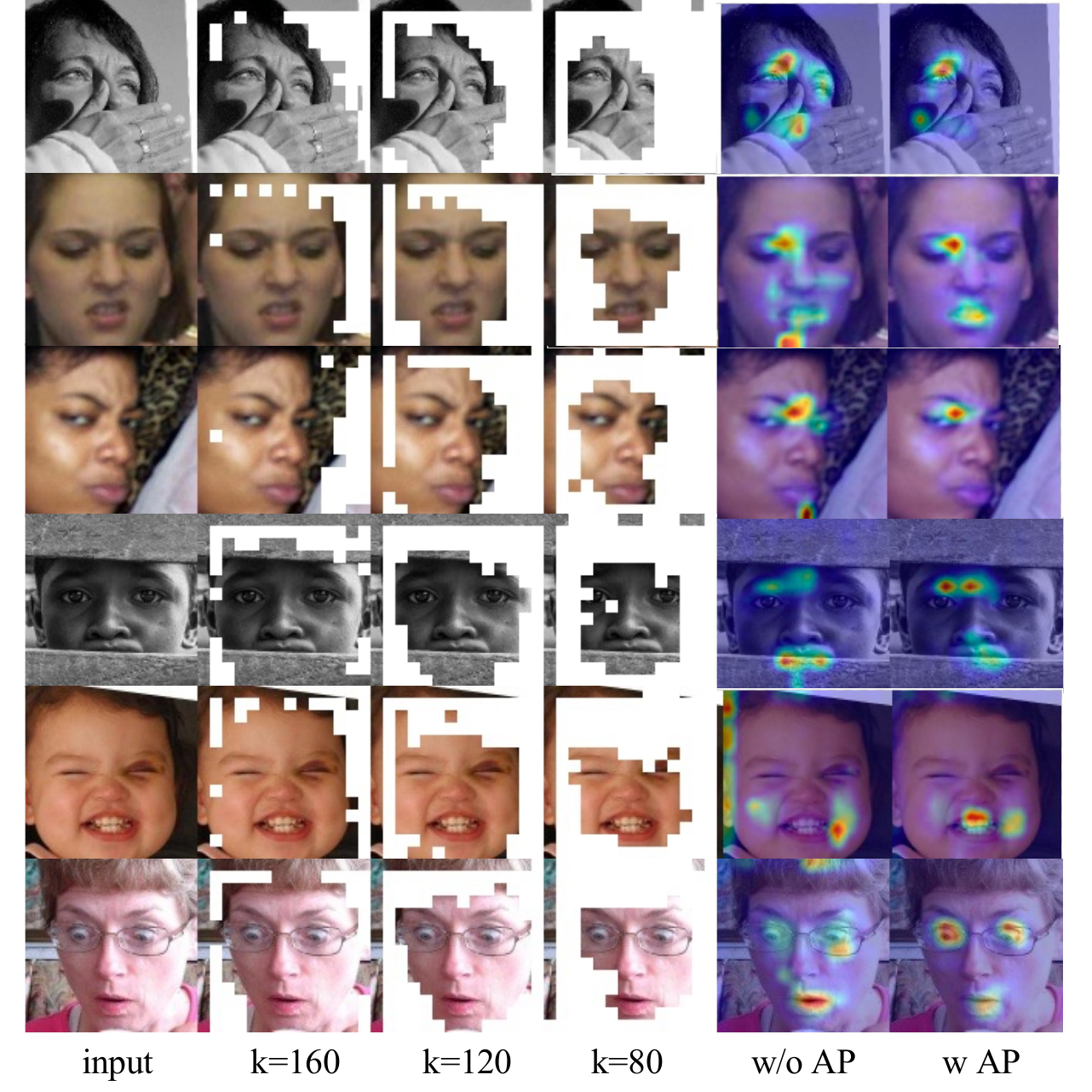}
\end{center}
\vspace{-3ex}
\caption{Visualization of attention maps in complex conditions. The second to fourth columns demonstrate attention maps in APP with different keep numbers of 160, 120, and 80, respectively. The last two columns are the attention maps in ATP without and with our proposed two AP modules, indicating the focused areas by Transformer.
}
\label{fig:vis}
\end{figure}

\textbf{Attention Visualization.}
In Fig.~\ref{fig:vis}, we visualize the attention map generated by our APViT on RAF-DB. The first column is the input images, and the second to fourth columns are reserved patches with our APP with different keep numbers. The pooled patches are erased to white background. The last two columns are the attention maps extracted by the method \cite{chefer2021transformer} to demonstrate the focus area of the model without and with the proposed AP modules. The selected images have complex conditions with occlusion and pose variation, varying with race and age. We can find that these attention maps are robust and accurate, and the pooled areas are noisy for the FER task. Thus 80 patches are enough to represent the facial expression. Compared with the baseline (w/o AP), AP modules could guide the model to pool the occluded and background areas and only focus on the distinguished ones,  \emph{e.g.} the hand in the first row and the chin in the second row. 
These visualizations also indicate that our approach is meaningful theoretically.

\section{Conclusion}
In this paper, we have proposed two Attentive Pooling (AP) modules with Vision Transformer (APViT) to utilize the pre-trained Transformer model for the limited size of the Facial Expression Recognition dataset. The proposed APViT could focus on the most discriminative features and discard the less relevant ones.
This could prevent the model from focusing on occlusion or other noisy areas. Experiments on six major in-the-wild FER datasets demonstrated that our APViT model outperforms the state-of-the-art methods. Visualization showed the intuition and robustness of our attentive poolings. 
We believe that decreasing the number of tokens is a promising direction to pursue. We hope our work can further inspire researchers to explore this paradigm with more effective methods.

\bibliographystyle{IEEEtran}
\bibliography{egbib}


%




\end{document}